# Offline Extraction of Indic Regional Language from Natural Scene Image using Text Segmentation and Deep Convolutional Sequence

[1]Sauradip Nag, [1]Pallab Kumar Ganguly, [1]Sumit Roy, [1]Sourab Jha, [1]Krishna Bose , [1]Abhishek Jha  and [2]Kousik Dasgupta

[1]Kalyani Government Engineering College, Kalyani, Nadia, India
[2]Faculty of Computer Science and Engineering, [1]Kalyani Government Engineering College, Kalyani, Nadia, India

sauradipnag95@gmail.com, pallabkumarganguly@gmail.com , sroy8091@gmail.com , sourab.jha@kgec.edu.in , krishna.bose02@gmail.com , j7.abhi@gmail.com , kousik.dasgupta@gmail.com

**Abstract :** Regional language extraction from a natural scene image is always a challenging proposition due to its dependence on the text information extracted from Image. Text Extraction on the other hand varies on different lighting condition, arbitrary orientation, inadequate text information, heavy background influence over text and change of text appearance. This paper presents a novel unified method for tackling the above challenges. The proposed work uses an image correction and segmentation technique on the existing Text Detection Pipeline an Efficient and Accurate Scene Text Detector (EAST). EAST uses standard PVAnet architecture to select features and non maximal suppression to detect text from image. Text recognition is done using combined architecture of MaxOut convolution neural network (CNN) and Bidirectional long short term memory (LSTM) network. After recognizing text using the Deep Learning based approach, the native Languages are translated to English and tokenized using standard Text Tokenizers. The tokens that very likely represent a location is used to find the Global Positioning System (GPS) coordinates of the location and subsequently the regional languages spoken in that location is extracted. The proposed method is tested on a self generated dataset collected from Government of India dataset and experimented on Standard Dataset to evaluate the performance of the proposed technique. Comparative study with a few state-of-the-art methods on text detection, recognition and extraction of regional language from images shows that the proposed method outperforms the existing methods.

**Keywords:** Segmentation , Text Recognition, Language extraction , Deep Learning , Bi-Directional LSTM , Tokenization, Neural Network.

## 1. Introduction

As internet and use of social media increases, the rise of easily accessible multimedia data, such as video, images and text created unprecedented information overflow. In contrast to this, information extraction from Natural Scene is still a daunting task and currently active area of research mainly because of the complexities faced during extraction. India being a vast country has a wide span of uncovered regions and various colloquial languages. Currently India has twenty six (26) different regional languages, many of these languages are native languages of mainly rural areas of India. So, extraction of these regional language from rural area is itself a challenging task, since the rural areas generally do not contain well defined address. Recent studies explored the extraction of Devanagari Scripts from text [8] but these do not give information about the specific regions where these languages are spoken, or the approximate location of the image. Modern applications focus on using metadata of an image to find the ground location of the image and from there find the accurate regional language. However, if images are captured using digital single-lens reflex (DSLR) or Digital Camera then the images may not be geotagged, so in such cases it will fail to get location. Therefore, it may be concluded that it is very much essential to extract location information from natural scene image offline, as it will be novel in its kind. Thus, the work in this paper focusses on extraction of location from natural scene image having address information offline and subsequently finding regional language of that region. This is interesting since we are using a text candidate segmentation, which outperforms the existing methods in performance in both low and high resolution. We are doing the entire process offline by training the proposed text recognition model for different regional languages. A sample collection of images has been generated from the Government of India dataset [14] for experimental study. Some sample images are shown in Fig. 1. The main contribution of proposed work is

- Using image correction for low quality images and using existing deep learning method to improve performance.

- Using deep convolutional sequence to detect Indic languages as shown in Fig 1.
- Detecting Location from a Image having Address Information in native languages
- Detecting the regional languages from its address information.

**Fig. 1.** Represents Sample Images in Different Languages having Address Information

The rest of the paper is laid out as follows: In Section 2, we discuss some related works regarding detection of text and text recognition using both traditional and Deep Learning based methods and finally extraction of language from it. In Section 3, the proposed text segmentation and connectionist text proposal network for Indic languages have been described alongside regional language extraction from location information. In Section 4, we discuss the details of dataset, the proposed experimental setup, comparison of proposed and existing methods and results in details. Section 5 points out the efficacy and drawbacks of the proposed approach, some direction towards future work and concluding with suggestive applications.

## 2. Related Work

Since natural scene image is occluded by various lightning condition, text orientation, inappropriate text information, so extraction of text candidate region from the natural scene image is very important in this literature. Tian et al. [1] proposed detecting text in natural image with connectionist text proposal network. The method detects a text line in a sequence of fine-scale text proposal directly convolutional feature maps. This method explores context information of the image for successful text detection. The method is reliably on multi-scale, multi-language text. Shi et al. [4] proposed detecting oriented text in natural images by linking segments. The idea here is that the text is decomposed into segments and links. A segment is an oriented text box covering part of text line while links connects two adjacent segments, indicating that they belong to the same text line. To achieve this, the method explore deep learning concept. Zhou et al. [5] proposed an efficient and accurate scene text detector. The method predicts text line without detecting words or characters based on deep learning concepts, which involves designing the loss of functions and neural network architecture. The method works well for arbitrary orientations. He et al [18] proposed Deep Text Recurrent Network (DTRN), which generates ordered high level sequence from a word image using Maxed out CNN and then uses Bi-LSTM to detect the generated CNN sequence. These works on text detection on natural scene images considers high-resolution, high-contrast images. These methods directly or indirectly explore connected component analysis and characteristics at character level for successful text detection of arbitrary orientation, multi-language, font face, font size etc. However, the performance of the methods degrades when the image contains both high and low contrast text. To overcome the problem of low contrast and resolution, the proposed method is effective as the Image with poor quality is fixed using Image Correction techniques and then uses EAST[5] to detect Text Boxes .It also uses convex hull to generate orientation wise correct text segmentation. The work in this paper presents a new method of text candidate region selection and segmentation by exploring. This work explores the problem of regional language detection from both low quality and high quality images. Since Deep Learning based approaches perform better than conventional approaches, so Neural Network based Architecture is used for both Text Detection and Text Recognition which is elaborated in details in the upcoming sections.

## 3. The Proposed Methodology

In this work a RGB image having at least 1 line of address Information is taken as input. This input image is then preprocessed for image correction using techniques like applying Gaussian Blur to reduce noise and applying Gamma Filter for enhancing contrast for night images. Then the Meta information is checked for GPS coordinates , if it is present then we proceed to find Regional Language.If not present we pass this corrected Image into EAST [5] Text Detction Pipeline which uses standard PVAnet and Non Maximum Supression Layer as a part of its Architecture. This returns the Bounding Box over probable Text Region . We binarize the image and perform periodic growing of the boxes on both sides until neighboring boxes overlap. In this way we make sure that the Adjacent Information is not lost . Still due to orientation some important information may be lost , so for that we apply Polygonal Convex Hull to correct the Segmentation and subsequently mask out the Color Image . After Segmenting the Text ROI from the Image, we use DTRN [18] since Address information can be in any font, we chose a method that is not dependent on Nature of Text. Motivated by the work of He et al [18] we use MaxOut CNN to get High Level Features for Input Text instead of grueling task of Character Segmentation from input since it is not Robust and Trustable. After this step as given in Figure 2, these high-level features are passed into Recurrent Neural Network architecture .Since we need address information which is quite sensitive to the context of a line , it may suffer from vanishing gradient problem . So Bi-LSTM is used since it solves vanishing gradient problem and it can learn meaningful long-range interdependencies which is useful in our context especially when our main task is to retrieve Address Information from Text Sequence. A Connectionist Temporal Classification (CTC) Layer has been added to the output of LSTM Layers for sequential labelling. We have trained this architecture for Hindi and Telugu which is our contribution. We pass the same input image for all the 3 language Architecture (English, Hindi, Telugu). The softmax Layer which is added in the End of MaxOut CNN Layers returns confidence scores for detection of High Level Features. We pass high level features of that Architecture whose softmax score is above some threshold. After recognition, we detect the language using Google's Offline Language Detector Module [28] and subsequently translate to English for better recognition rate.

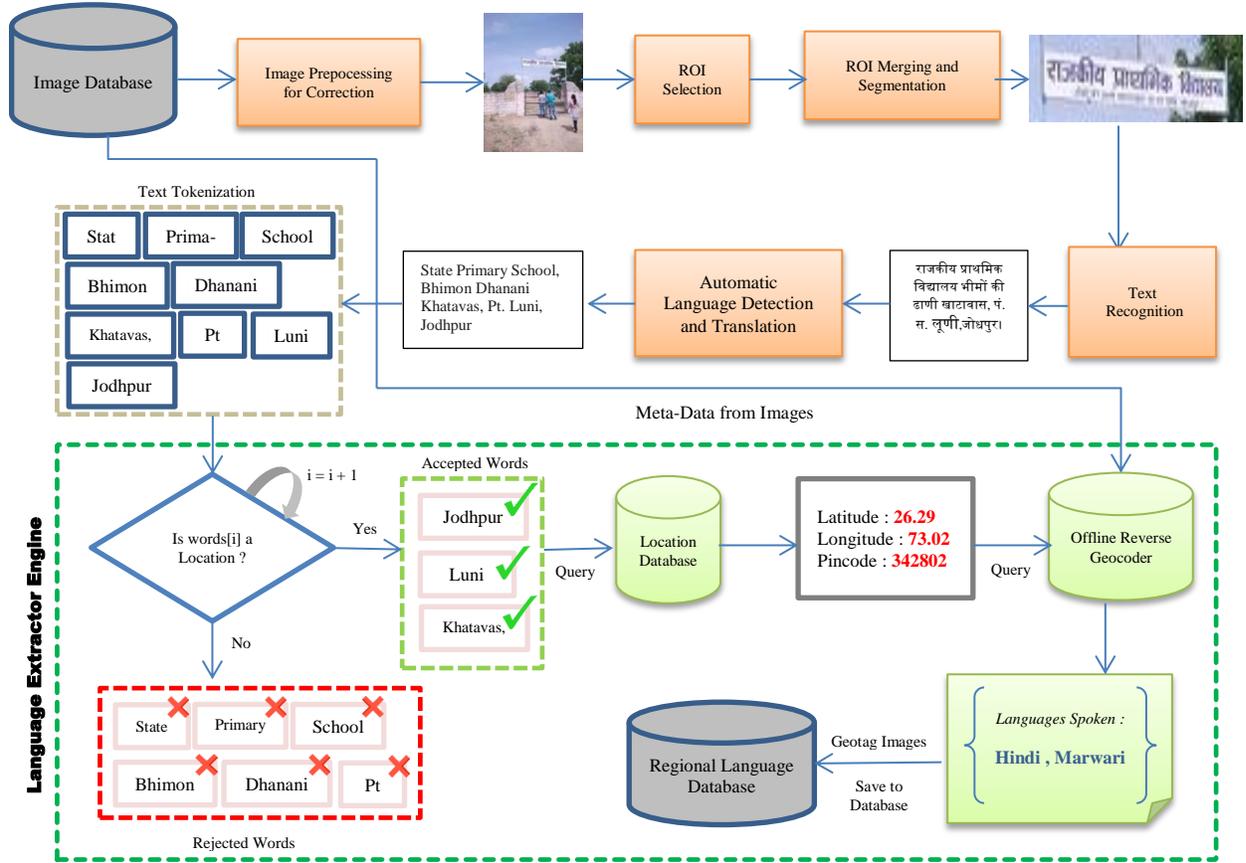

**Fig. 2.**: Framework of the Proposed Method

The proposed Approach then uses Moses Tokenizer [19] to tokenize the text as this tokenizer tokenizes the words without punctuation from raw translated text. Each of these tokens/words are queried individually on Post Office Database collected from Gov-ernment Of India to obtain the GPS coordinates of all the common post office of the locations found using Tokenization. After obtaining (Pincode, Longitude, and Latitude) for a particular image, we use reverse geocoding [29] and database RLDB created by collecting Regional Language Information found in [13]. From this, we obtain the Regional Language Spoken for the particular location found in Input Image and finally we Geotag the Images and save them. The complete framework of the proposed method can be seen in Fig 2.

### 3.1 Image Correction

Any picture from natural scene can contain certain amount of noise, blur and varying lightning condition, which may affect any method for text candidate selection. So extra care must be taken to ensure that the performance must not drop irrespective of the quality of picture if it has at least one address information. So, to correct the Image before text detection , the input image is preprocessed. Bright or Dark Image is identified using Histogram of Intensity.If the input Image is Night Image(dark) then a Gamma Filter with Gamma Value 2.5 is applied. The value 2.5 is empirical and is obtained after proper experimentation. So, this gamma value is used to correct it as shown in Fig 3(b). Noise is removed from color image by using non-local de-noising method [9] and unsupervised Weiner Filter [10] is used to de-blur the de-noised image as shown in Fig 3(c).

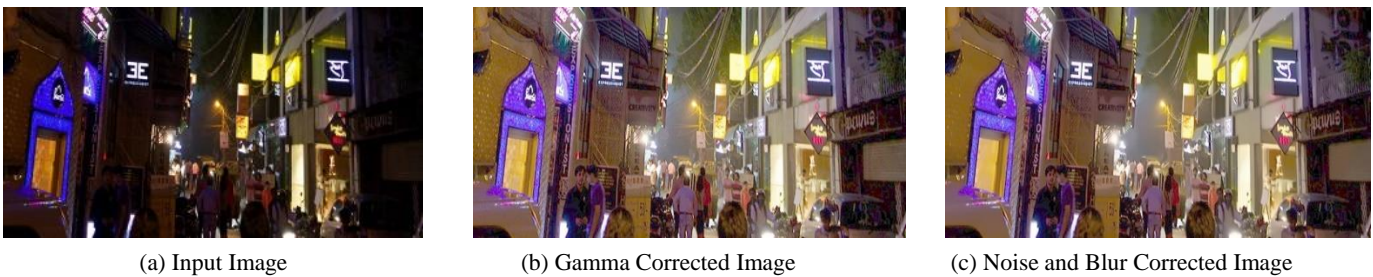

(a) Input Image    (b) Gamma Corrected Image    (c) Noise and Blur Corrected Image

**Fig. 3.** Depicts how Input Image is preprocessed before Text Detection

### 3.2 Text Detection and Candidate Segmentation

Text Detection in Natural Scene has been an area of research for some years in the Recent Past . Conventional Approaches Rely on manually designed features like Stroke Width Transform (SWT) , MSER based methods which identifies character candidates using Edge Detection or extremal region Extraction or contour features .Deep Learning Methods outperform these methods as these methods fall behind in terms of robustness and accuracy .

**Text Detection :** Since Zhou et al. [5] proposed EAST which predicts text line without detecting words or characters based on deep learning concepts , this approach is suitable for Extracting Information like Text in Multiple Fonts . EAST[5] also implemented Geometry Map Generation which can Detect Text in Arbitrary Orientation without losing Text information with the help of rotating box (RBOX ) and Quadrangle (QUAD). But EAST[5] fails for Low Quality Images since Trained Model used High Quality Images from Standard Datasets . As mentioned in Section 3.1 , the Low Quality images are improved and fed to the Text Detection Pipeline . The pipe line proposed by Zhou et al. [5] consists of only 2 stages namely Fully Convolutional Neural Network (FCN) and Non-Maximum Suppression ( NMS ) . The FCN internally uses a Standard Neural Network Architecture called PVAnet which consist of 3 internal Layers : Feature Extraction Layer , Feature Merging Layer and Geometry Map Generation for Bounding Box which is after the Output Layer as depicted in Fig 4 .The input image to this step is the Uncorrected Low/High Quality Images . The Input image is then resized and fed to the Neural Network Pipeline where it extracts the Feature relevant to text in $1^{st}$ Layer , $2^{nd}$ Layer it merges the features and extract possible text region mapping using Geometry Map Generation losses . After This step the NMS Layer selects the maps which are only for text and marks a Bounded Box around it as shown in Fig 4(c).

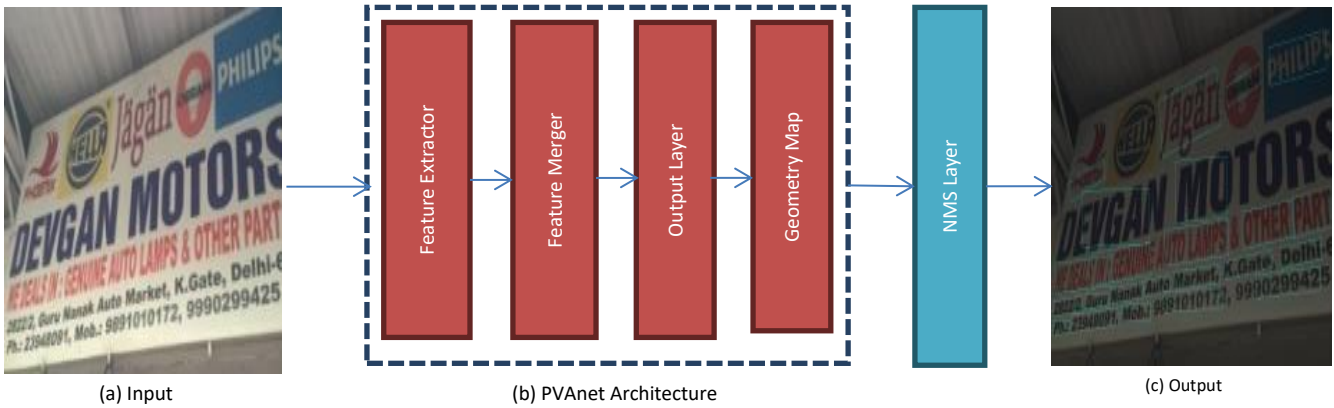

**Fig. 4.** Architecture of Text Detection Pipeline of EAST[5]

**Text Segmentation :** Since the Architecture mentioned Above selects individual Boxes it may be difficult to extract location information . Since each location connected by '-'(dash) may represent a location and EAST alone will only detect the two connected words individually which may create ambiguity in location extraction . Hence we need to process the Image and segment out the Text Candidate Region. The Proposed Pipeline thus contains 3 key steps : Image Correction , Text Detection , Merging and Segmentation as shown in Fig 5 . Input Image is corrected by maintaining Intensity , Blur Correction and Denoising . This image is then passed onto the Existing EAST[5] Pipeline as mentioned in Fig 4 . This FCN takes corrected images as input and outputs dense per pixels prediction of Text Lines . This eliminates the time consuming Intermediate Steps which makes it a fast text detector. The post processing steps include thresholding and NMS on predicted Geometric Shapes . NMS is a post processing algorithm responsible for merging all detections that belong to the same object . The output of this stage is bounded Box on the text regions as shown in Fig 5(d).

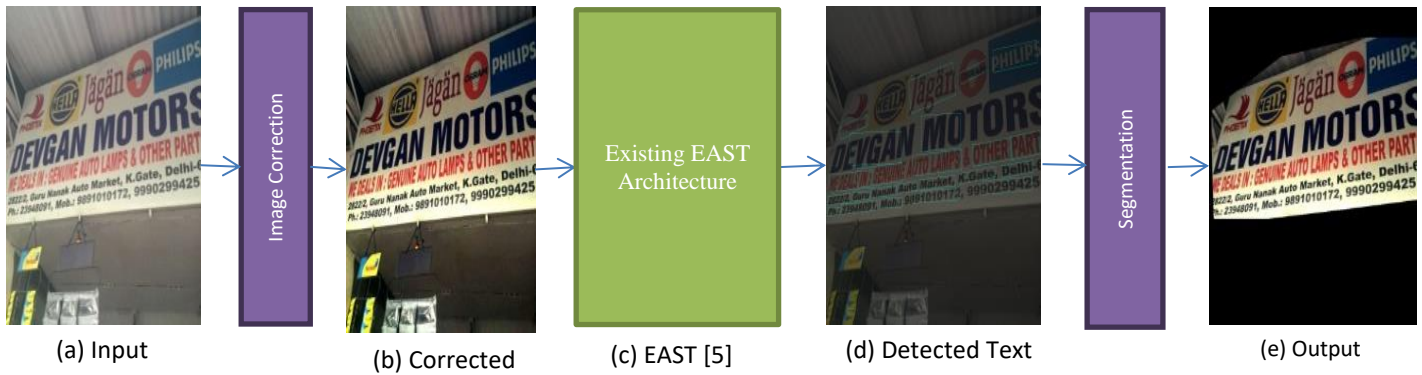

**Fig. 5.** Architecture of Proposed Text Detection Pipeline

After Formation of Bounding Boxes over the Image as shown in Fig 6(c) the box sizes are increased in a periodic manner on both the Sides until it overlaps with neighboring bounding box . In this way , we make sure that the Neighboring Textual Information is not lost . This is represented in Fig 6(d). After this operation, we observe that the segmented image is not uniform (in Fig 6.(d) ) and has lost Information which may affect the overall results since address information is interlinked and order of text matters. So the proposed approach first binarizes the incorrect segmentation and then uses Polygon Convex Hull as in Fig 6.(f) to correct the Image and the text information which was lost due to orientation , and box merging are recovered . Finally, we masked out the region from Color Input Image as shown in Fig 6(g) and Fig 5 (e) . Now this will be the input to the Text Recognition Stage .

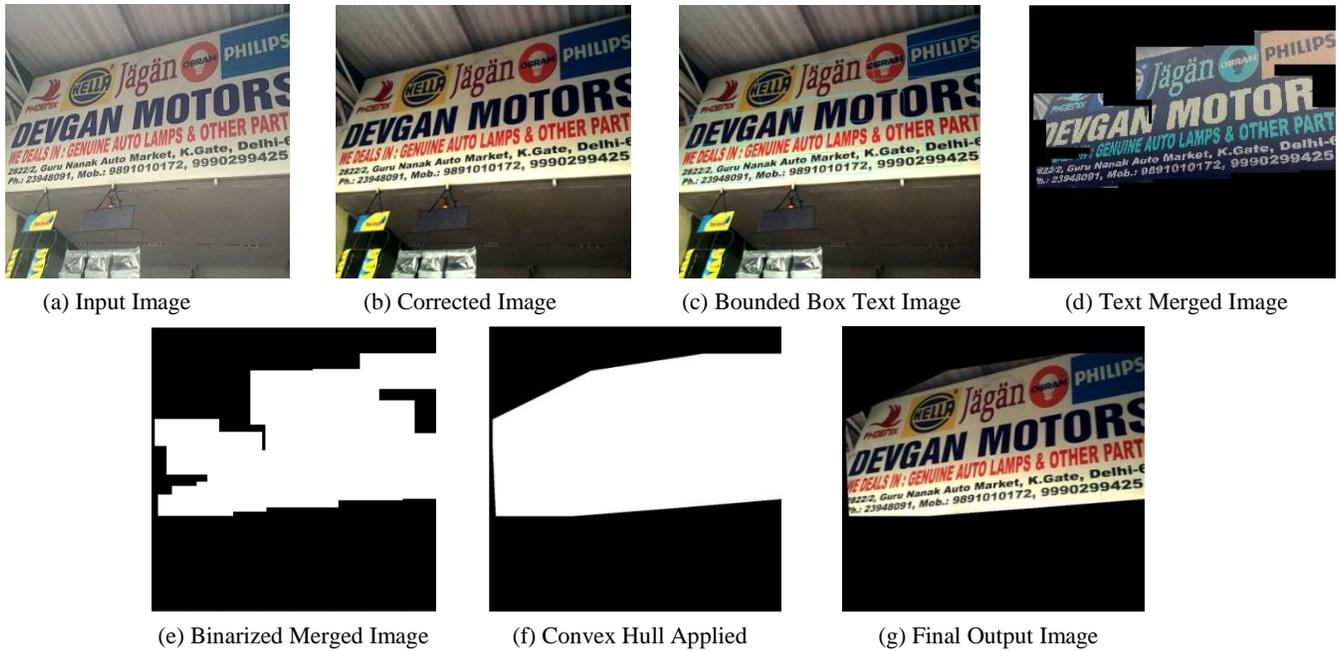

**Fig. 6.** Represents Step by Step Process of Detecting Text from Input Image and Segmenting it for Text Recognition Step

Now the Text Recognition Method will not search the entire Image for text instead it will search on a smaller region which makes this method quite Fast and Accurate . The whole process of Execution takes less than 15 seconds for images having size of 2000 x 1500 on a Intel Core i5-5200U CPU @ 2.20GHz Processor with 8GB Ram . Hence it segments out the Text in a time which is almost equivalent to the time taken by EAST[5] to process a image individually .

### 3.3 Text Recognition and Key Word Extraction

Detecting Text from Natural Scene is a daunting task due to the changing background in different images and other external factors. Several works have been done on detecting text from Image using both traditional methods and more recently with Deep learning methods. But in most cases the Deep Learning methods outperforms the traditional methods due to the complexity and orientation. However, it must be noted that the images used for training and testing the Deep Learning Based Methods require a high DPI, so images must not be of low resolution. The modern approaches use Deep Learning Based Text Detection and Text Recognition Pipeline [5]. However, these methods fail to detect a candidate region in low resolution images. But the proposed approach can detect text candidate in both high and low-resolution images better than other existing approaches and for text recognition purpose we have used Deep Learning Based Text Recognition Method as proposed by Tian et al. [1]. Since the method with connectionist text proposal network detects a text line in a sequence of fine-scale text proposal directly convolutional feature maps. Unlike the problem of OCR, scene text recognition required more robust features to yield results comparable to the transcription-based solutions for OCR. A novel approach combining the robust convolutional features and transcription abilities of recurrent neural networks as introduced by [37]. DTRN [18] uses a special CNN called MaxOut CNN to create High Level Features and RNN is used to decode the Features into word Strings. Since maxout CNN is best suited for character classification we use this on entire word image instead of whole image approach. We pass the bounding box text information of the boxes which lie in the Segmented Text Region shown in Fig 6(g) to this MaxoutCNN for extracting High Level Features for Extraction . The Features are in the form of sequence $\{x_1, x_2, x_3, \ldots, x_n\}$. These CNN Features $\{x_1, x_2, x_3, \ldots, x_n\}$ are fed to the Bi-Directional LSTM in each time sequence , and produces the output $p = \{p_1, p_2, p_3, \ldots, p_n\}$. Now the length of input sequence X and length of output sequence p is different , hence a CTC Layer is used which follows the equation :

$$S_w \approx \beta(\arg\max_\pi P(\pi|p)) \qquad (1)$$

Where $\beta$ is projection which removes repeated labels and non-character labels and $\pi$ is the approximate optimized path with maximum probability among LSTM outputs . For example , $\beta(--ddd-ee-l-hh---i-) = delhi$, hence it produces sequences $S_w = \{s_1, s_2, s_3, \ldots, s_n\}$ which are of same length as Ground Truth Input . India being a vast Country has a diverse spread of Language [13], since training data collection for these languages are challenging, for the present we have focused to use

English, Hindi and Telegu Text Recognition in this work.  For training the CNN part of DTRN [18] we collected individual character Images from various datasets [14-17] and we trained the Maxout CNN individually for English, Hindi and Telegu. For training samples, we collected individual character from [15], [16], [17] for English Texts, Hindi and Tamil were obtained from [14], [17] and publicly available images in Google. In total we trained characters of multiple fonts, each class has 10 different sample of varying width size and style which were collected from around 2000 Images for English, 1100 for Telegu and 1800 for Hindi Scripts. For training RNN part, we took the images from the training samples [14-17] and cropped out word images individually and collected 3700 words for English, 1900 for Telegu and 2900 for Hindi Language. Although, He et. al. [18] proposed DTRN for word recognition, we have extended this to sentence recognition with good accuracy. The Hyperparameters used are :  Learning Rate fixed to 0.0001 and momentum of 0.9. The Loss Function used in this Recurrent Layers is defined in Equation(2) as shown below :

$$L(X, S_w) \cong - \sum_{t=1}^{T} log\big(P(\pi_t|X)\big) \qquad (2)$$

where $\pi \in \beta^{-1}(S_w)$ . Then this approximate error is backpropagated to update parameters . The Raw Text is then Translated from Native Language to English Language using [12]. After this step, the proposed approach uses Moses Tokenizer [19] which is especially good in terms of Tokenizing phrases/words without Punctuation than other existing Tokenizers. Each Token is then Passed to Google Location Offline Database to check if it is a location, if it is location, we categorize such Tokens as KeyWord which are most likely to generate Location Information else we reject the token.  The reason behind Tokenizing text is to remove ambiguity for name of place which can exist in two or more different states whose regional language is different. For example, "Raniganj Bazar" is a place in the State Uttar Pradesh, India but "Raniganj" is also a Place in state West Bengal, India where Regional Languages are Hindi and Bengali respectively. In such cases, we look for secondary information which is discussed later in Section 3.4

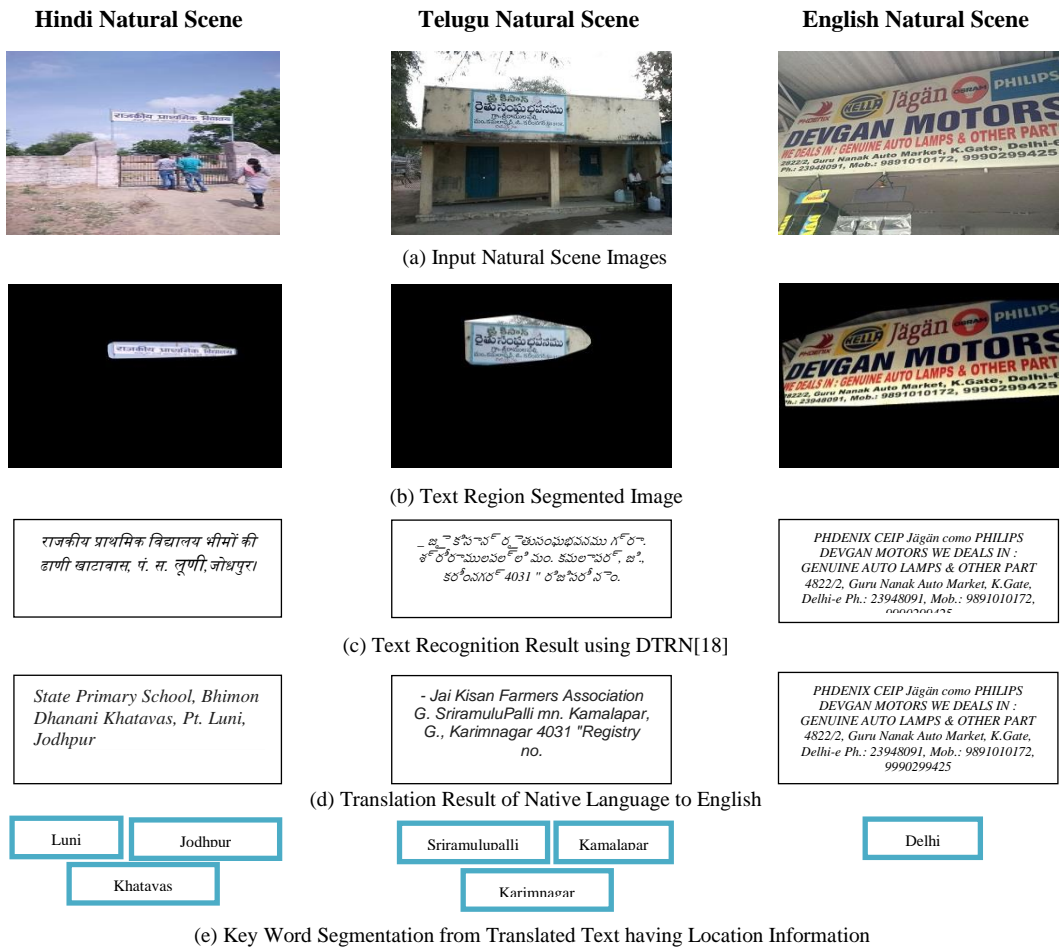

**Fig. 7.** Representation of how Key Location Word is Extracted from Segmented Text using DTRN

The illustration in Fig 7 explains the importance of Text Recognition in key word extraction from the Segmented Image . So segmentation and text recognition plays an important role in detection of Regional Language . Moses Tokenization is particularly Important in Conversion of Fig 7(d) to Fig 7(e) since any wrong key word may result in extraction of Different Regional Language . The step by step process from the bounding box information of the Input Image to Key Word Extraction including the intermediate steps are clearly illustrated in Fig 8

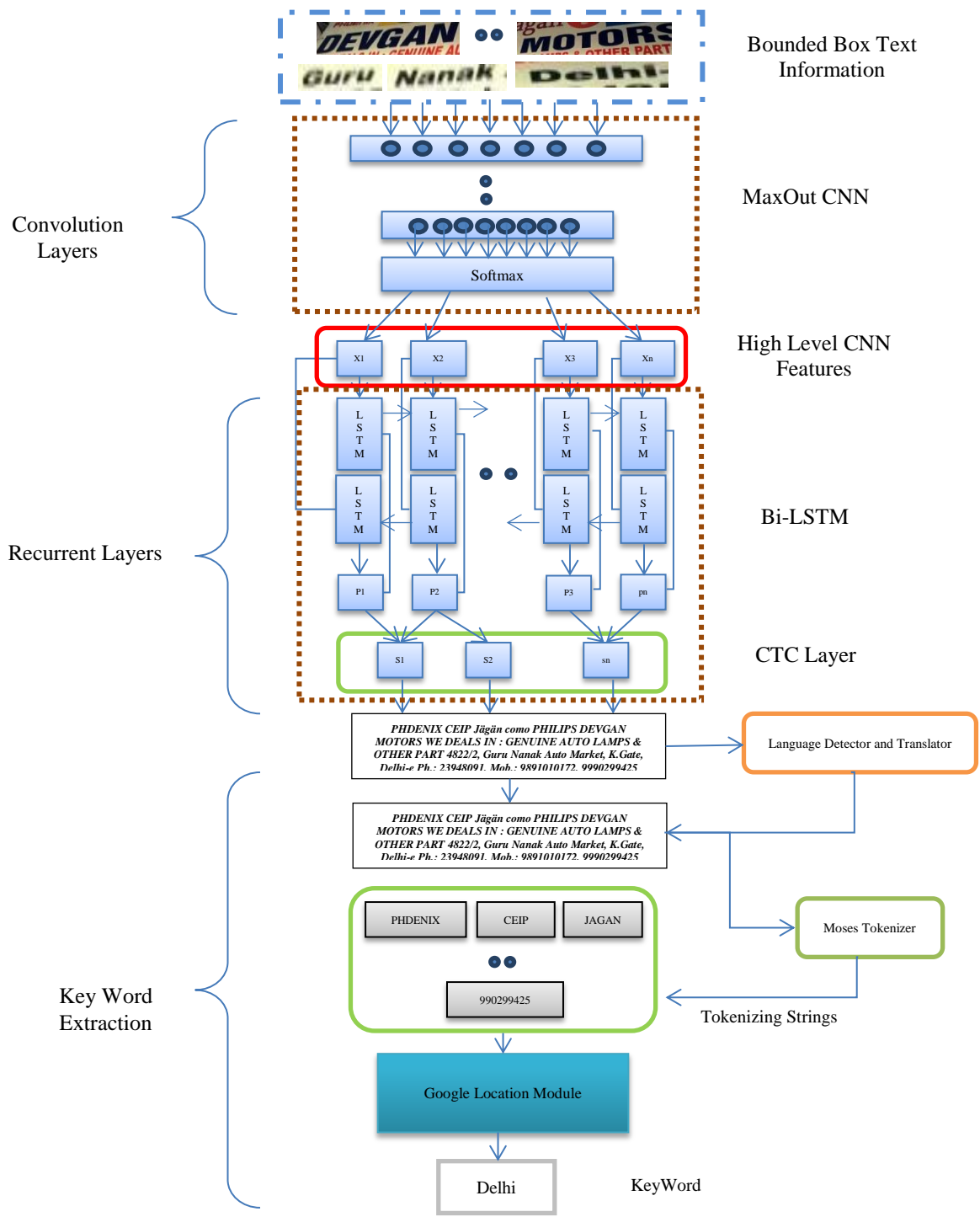

**Fig. 8.** Pictorial Representation of Text Recognition and KeyWord Extraction Pipeline Architecture

**3.4 Regional Language Extraction from Location Information**

The key words obtained from previous section is used for detecting regional language. For each Token we query the Database collected from Post Office Database from Government of India. Our choice of database is Post Office because Post Office is present in every corner of India and since we are targeting regional language we are covering entire Country. The key words are queried in the Database where the Primary Key is Composite Key of (Division Name, Taluk, Pin code), Taluk is generally a Administrative centre controlling several villages, hence along with Pin code it determines a region uniquely. This Database is referred to as CSDB (City and States Database) [20] as shown in Table 1. Since we need location information, so we also need another database which contains GPS information corresponding to such regions. For that we consider the Database [21] which contains City ID, City, State, Longitude and Latitude as attributes and City Id as Primary Key. This database is referred to as LLDB (Latitude and Longitude Database) as shown in Table 2.

**Table 1.** Representing the Post Office Database [20] CSDB with its Schema and Internal Structure

| Div_Name | Pincode | Taluk | Circle | Region | District | State |
|---|---|---|---|---|---|---|
| A-N Islands | 744112 | Port Blair | West Bengal | Calcutta HQ | South Andaman | Andaman and Nicobar Islands |
| Saharsa | 852201 | Kahara | Bihar | Bhagalpur | Saharsa | Bihar |

**Table 2.** Representing the GPS Database [21] LLDB with its Schema and Internal Structure

| City Id | City_Name | Latitude | Longitude | State |
|---|---|---|---|---|
| 1 | Port Blair | 11.67 N' | 92.76' | Andaman and Nicobar Islands |
| 255 | Saharsa | 25.88 N' | 86.59 E' | Bihar |

Using a Nested Query as shown below in Fig 9 we will fetch all possible tuples of combination (Latitude, Longitude, Pin Code) from the Key Word Token individually for a particular Image and then we take the common tuple among all key words of the Image to get the approximate location. In cases where we have ambiguity in Location as mentioned in previous Section 3.3, we first observe the common tuple among Key words, if not found then we conclude that there is ambiguity and we are missing some information. To solve this issue we propose a method to check secondary information like the detected language of Text Recognition, even if we cannot separate after this step then we combine adjacent tokens pairwise from left to right and re run the same queries again to check common tuple. If we fail after this step, we conclude that Location cannot be found and some more operations or information needs to obtained. After this operation we perform reverse Geocoding [29] on (Longitude, Latitude) as input to obtain the approximate location. Subsequently we query the location from Language Database (RLDB), which is created by us with data collected from Wikipedia. Finally, we obtain the Regional Language set which is obtained for a particular input image. We Geotag the Image with the Language Information and store it in a Database.

```
for i in keywords :
    tuples[j++] = sql.execute(" select t1.latitude,t1.longitude,t2.pincode
                from LLDB t1 ,CSDB t2 on t2.Division_Name = t1.City_Name
                where t2.Taluk = '"+ i + "';  ")
end
    com_loc=common(tuples);
if (com_loc.size() > 0 )
    place = reverse_geocode(com_loc.latitude,com_loc.longitude)
else
    Solve Location Ambiguity
end
```

**Fig. 9.** Illustrates Pseudo Code for Extraction of Location from Keywords

## 4. Experimental Results

For Evaluating the Proposed Method we created a Complex Dataset comprising of Standard Datasets like ICDAR-15 , KAIST , NEOCR [14-17] . The input images contain Languages of three Classes namely English, Hindi and Telegu. Each class has images ranging from daylight image to night image. Images were reportedly taken from good quality cameras, but to make the dataset diverse, images captured from Smartphone cameras of both high and low resolutions are added. The Total Size of Datasets are 8500 images out of which 30% images are low resolution ( < 400 x 400 pixels) and rest are high resolution. Out of these 3700 words for English, 1900 for Telegu and 2900 for Hindi Language. From this dataset well shuffled 70 % Data are used for training and remaining 30 % are used for testing the proposed CNN and RNN Models. To Measure the performance of Text recognition we use standard measures namely Recall (R) , Precision ( P ) as defined in [22] . To evaluate the Performance of Text Detection we use standard measures as defined in [23-24]. We also compare location accuracy with Google Maps to justify our offline location searching. To prove usefulness of method in Text detection, we consider standard Deep Learning Methods, EAST[5], Tian et al [1] and Shi et al. [4]. As per our knowledge, these methods work best for daylight, good resolution images. For effectiveness of our approach and fair comparative Study, we pass the input images from the Database as it is including night images, low resolution, high resolution images. For text recognition, we will compare our method with Traditional Tesseract OCR [23], and OCRopus [24]. We pass the Input Images Directly to OCR [23-24] for fair comparison. So text detection, Text recognition are the key steps of this method. We also evaluate the Accuracy of Location found out using proposed method .

The following Figure represents the sample Database considered for our literature for Testing the efficacy of the Proposed Method

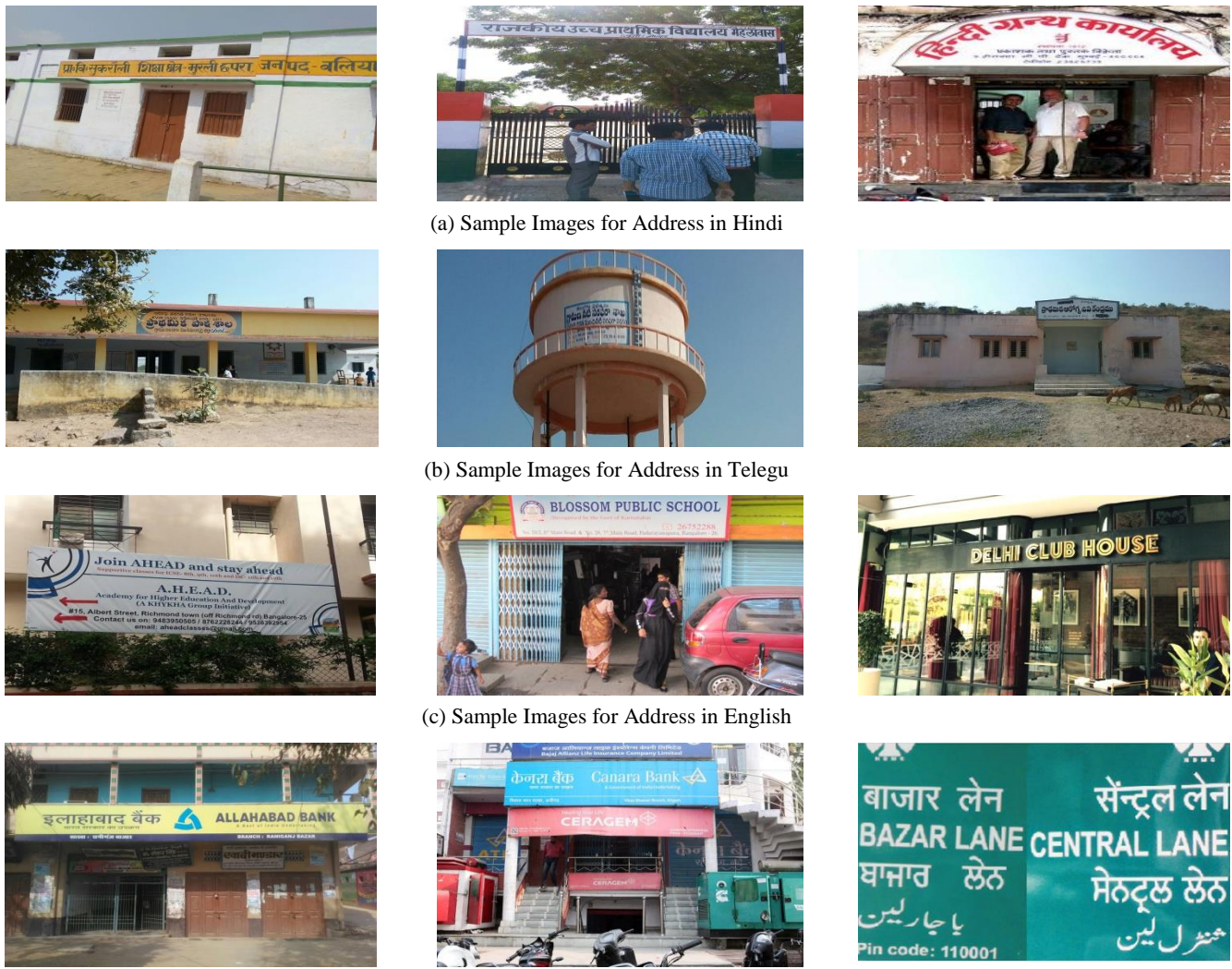

(a) Sample Images for Address in Hindi

(b) Sample Images for Address in Telegu

(c) Sample Images for Address in English

(d) Sample Images for Address in Mixed Indic Scripts

**Fig. 10.** Sanmple Images of the Databases

### 4.1 Evaluating Text Detection

Qualitative results of the proposed and existing methods for text detection in Natural Scene are shown in Fig 11. It is observed that the Proposed Method performs well in comparison to existing methods [4 ,5]. Although Deep Learning solves many complex problems like occlusion, font size, it requires good number of training samples, high quality Images. However, it performs badly in low light images as depicted in Fig 11.1. (a), however proposed approach corrects any Input Image using Gamma Filter as explained earlier, hence detection rate is high. The advantage of proposed method over existing method of text detection lies in the fact that it performs exceedingly well in low resolution and low light images.

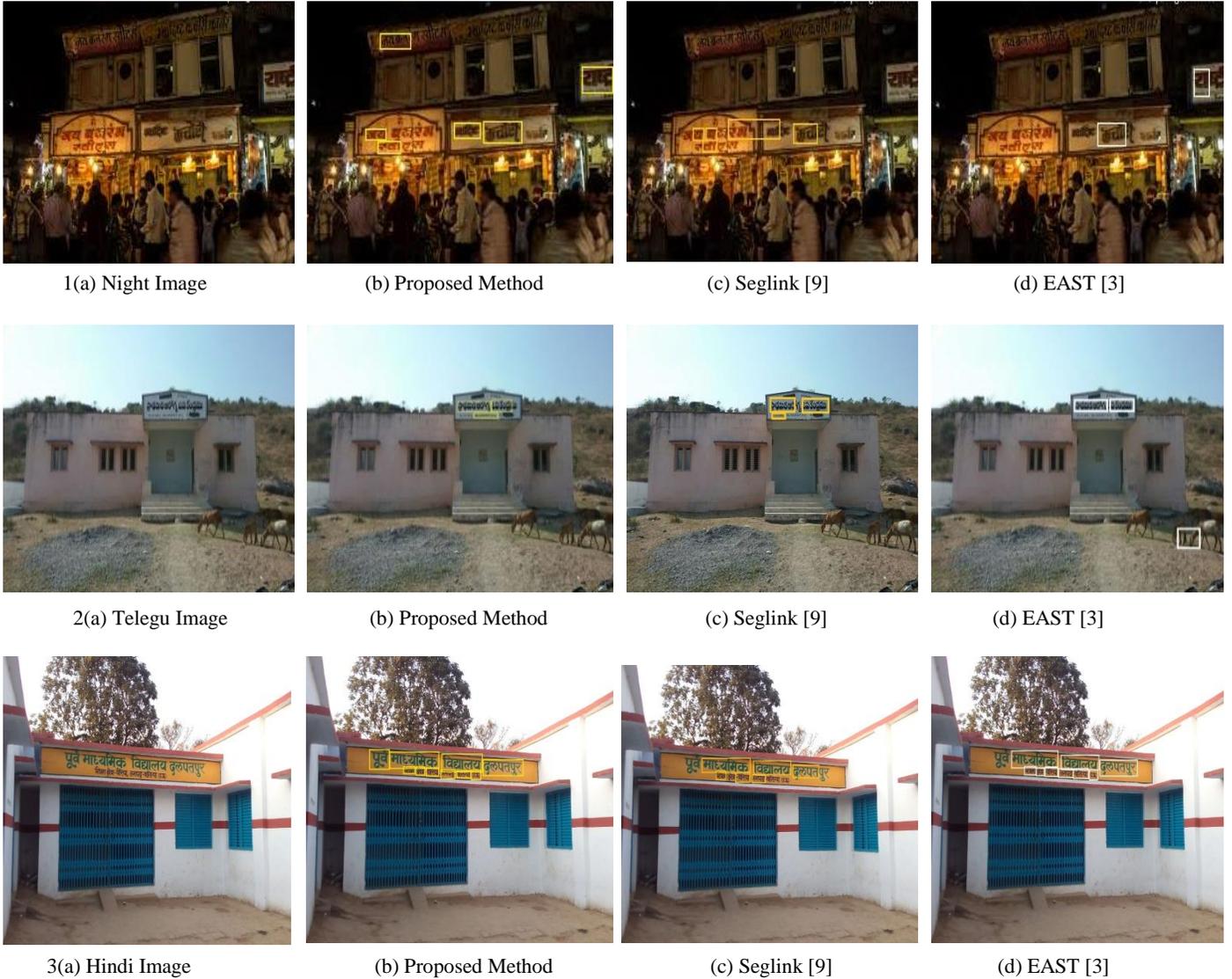

1(a) Night Image    (b) Proposed Method    (c) Seglink [9]    (d) EAST [3]

2(a) Telegu Image    (b) Proposed Method    (c) Seglink [9]    (d) EAST [3]

3(a) Hindi Image    (b) Proposed Method    (c) Seglink [9]    (d) EAST [3]

**Fig. 11.** Represents Text Detection on Image Database for Proposed Approach and Existing Approaches

Quantitative results of the proposed and existing methods for three databases are reported in Table 3 where it is noted that the proposed method is best at precision , recall , F-measure in comparison to the all the existing methods. It is obvious since, the Image Database (ICDAR-15 , KAIST , NEOCR [14-17] , Mobile Camera Images) contains images ranging from low resolution to high resolution of various fonts and occlusion, less text information, Deep Learning based approaches sometimes fails to detect text regions but with correction these images produce good results . EAST [5] scores 2nd best among the 3 deep learning based methods and Seglink [4] performs poorly .

**Table 3.** Performance of the proposed and existing methods for text detection

| Methods | Technique | Precision | Recall | F-Score |
|---|---|---|---|---|
| **Proposed Method** | Deep Learning | **0.81** | 0.78 | **0.80** |
| Seglink [4] | Deep Learning | 0.73 | 0.74 | 0.73 |
| EAST [5] | Deep Learning | 0.77 | 0.75 | 0.76 |
| Yin et. al [30] | Image Processing | 0.68 | **0.86** | 0.71 |

### 4.2 Evaluating Text Recognition

Text Recognition in Natural Scene is a very interesting work. Due to orientation, sometimes feature extraction becomes quite difficult. Traditional approaches like MSER Features work well in images having good lightning, high resolution Images but fail significantly in low resolution and Night Images. Preprocessing improves the Accuracy but not to desired standards. Since we are dealing with Location Extraction from Text, recognition of text is very important in this context, so partial text recognition is not reliable. Huang et al [25] proposed MSER based Text Recognition but it detects simple text orientations, complex text orientation does not give good results using MSER features. Roy et al. [26] proposed conditional random field model defined on potential character locations and the interactions between them. This works well for individual words and not on region-based word recognition, where words are interlinked, so missing on word leads to lossy information. Although many works used Tesseract OCR [23] for text recognition but since it was originally developed for Document Based Extraction, performance significantly drops in Natural Scene due to complex background. Similar scenario with OCRopus [24], it is also developed for document based extraction but it performs better than Tesseract OCR [23] as shown in Table 4 and Fig 12, since it uses preprocessing like Deskew , DeNoise , Demask before performing Text recognition. Alzaman et al [31] proposed a fast text recognition technique using a combination of label embedding and attribute learning. Zhao et al [32] proposed a text recognition pipeline where he proposed a modified VGG-Net architecture for character recognition .

So Deep Learning based method performs significantly better in such scenarios, although DTRN [18] works well for individual words, according to our experiments it performed well in Region based word recognition. DTRN [18] uses MaxOut CNN to extract High Level Features from Region of Interest and then Uses Recurrent Layers to connect all information sequentially. Hence it performs better than existing Approaches. The metrics used for performance calculation are Precision and Recall as mentioned earlier.

**Table 4.** Performance of the proposed and existing methods for text recognition

| Methods | Precision | Recall |
|---|---|---|
| **DTRN [18]** | **0.77** | **0.81** |
| Tess-OCR [23] | 0.32 | 0.46 |
| OCRopus [24] | 0.58 | 0.73 |
| Almazan et al. [31] | 0.67 | 0.71 |
| Zhao et al [32] | 0.73 | 0.66 |

| English Text | DTRN [18] | Tess-OCR [23] | OCRopus [24] |
|---|---|---|---|
| 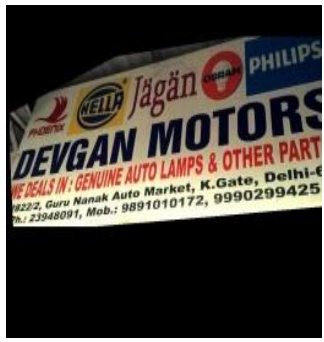 | 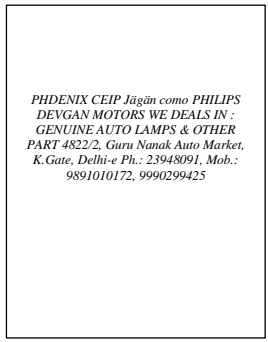*PHDENIX CEIP Jägän como PHILIPS DEVGAN MOTORS WE DEALS IN : GENUINE AUTO LAMPS & OTHER PART 4822/2, Guru Nanak Auto Market, K.Gate, Delhi-e Ph.: 23948091, Mob.: 9891010172, 9990299425* | 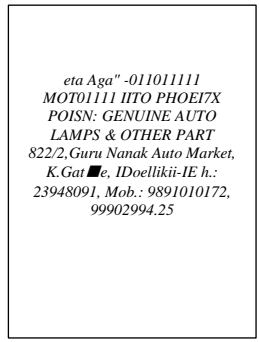*eta Aga" -011011111 MOT01111 IITO PHOEI7X POISN: GENUINE AUTO LAMPS & OTHER PART 822/2,Guru Nanak Auto Market, K.Gat■e, IDoellikii-IE h.: 23948091, Mob.: 9891010172, 99902994.25* | 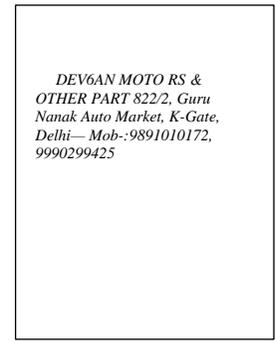*DEV6AN MOTO RS & OTHER PART 822/2, Guru Nanak Auto Market, K-Gate, Delhi— Mob-:9891010172, 9990299425* |

| Telegu Text | DTRN [18] | Tess-OCR [23] | OCRopus [24] |
|---|---|---|---|
| 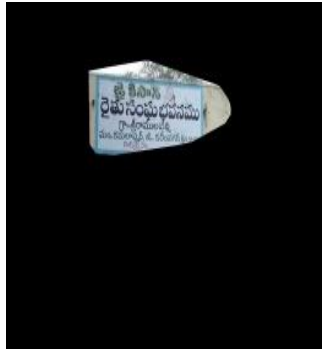 | 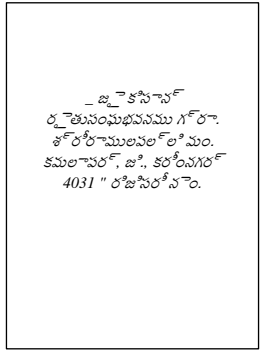 | 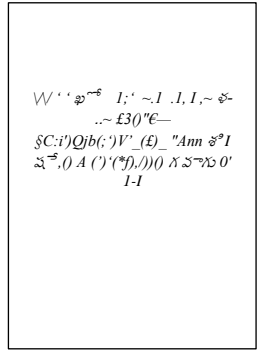 | 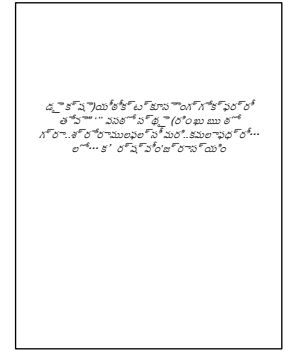 |

| Hindi Text | DTRN [18] | Tess-OCR [23] | OCRopus [24] |
|---|---|---|---|
| 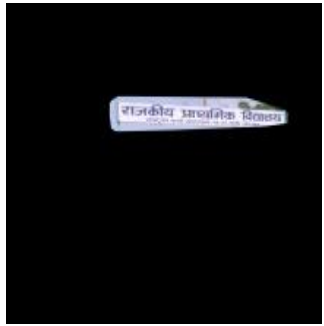 | 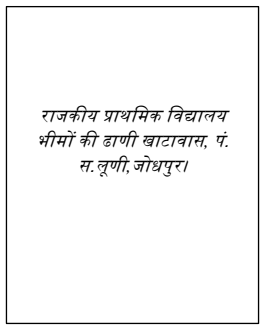 | 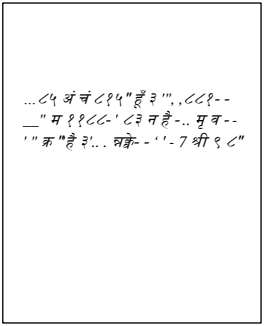 | 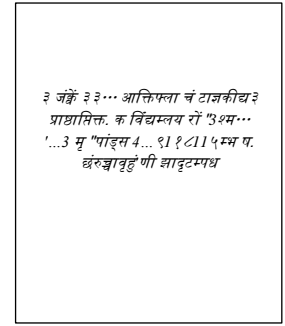 |

**Fig. 12.** Represents Text Recognition Results for Indic Scripts using both Proposed and two existing methods

Location Accuracy is also important since we are investigating Regional Language in this literature. Regional language in India however does not change within few kilometers distance until and unless the Location is close to an International Border or State Border, we consider those as exceptions. For Measuring Location, we consider the Key words as locations and search for their GPS coordinates, since this approach is offline there is a high chance of error. So as a ground truth we search Googles Online Maps with the Key words and compare with the Proposed Offline Version to calculate the Distance(d) between the Proposed Point and Ground Truth Point. We run each image individually on 8500 images and calculate 'd' and finally we take average, so we found out average distance of proposed and ground truth point as a metric to justify the Accuracy. In our Experiments, the Average Distance between Proposed Point and Ground Truth Point comes out as 11.87 Km as shown in Fig 13

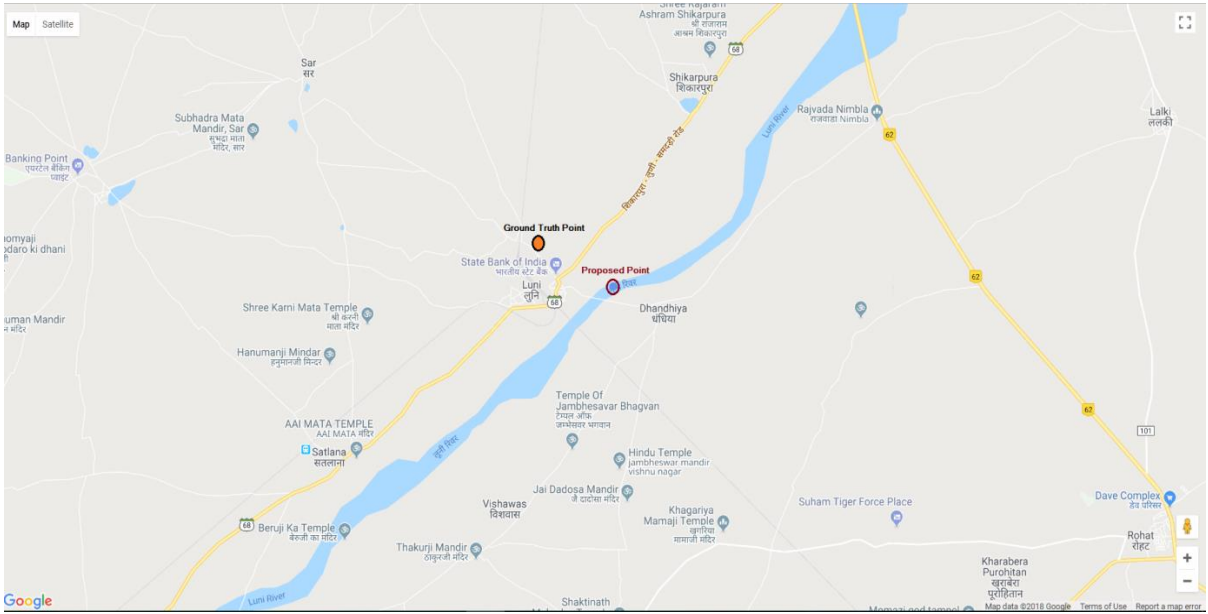

**Fig. 13.** Illustrates the Proposed Location obtained empirically from LLDB and Ground Truth Point obtained from Google Street on Indian Map for Location Luni , Jodhpur

## 5. Conclusion And Future Work

This work highlights a novel method in extracting regional language from an Image containing any Address information. This is a novel work of its kind where regional language is extracted from an Image without Internet (offline). The proposed method also defines a unique Text detection approach where Images of low resolution and varying font size can be handled easily and text can be extracted. However, it is noted that the proposed method is not robust to occlusion, severe blur and too small font as shown in Fig 14(c) while capturing image from long distance camera hence address cannot be extracted. The proposed method also fails in scenarios where Advertisements which contains location of office not in current state as represented in Fig 14(a) , here Picture Location is Kolkata but proposed method also captures New Delhi but the regional Language of New Delhi and Kolkata are different. It also fails in scenarios like Fig 14(b) where the First word is a country which has different regional language . Currently Text Recognition can detect English, Hindi and Telegu Scripts, in future the proposed work can extend to 26 Officially accepted languages in India for unearthing the rarest regional Languages which is unknown to mankind. The work needs to further address the location ambiguity using other information to detect accurate languages. The Query for selecting the Postoffice can also be optimized based on Ground truth data. The application of this literature lies in the field of identification of unknown places in India and also can be used in computer forensics .

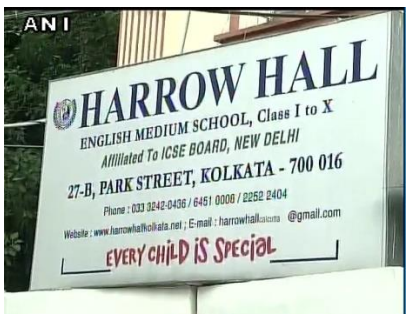 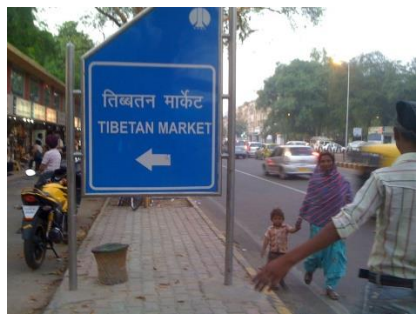 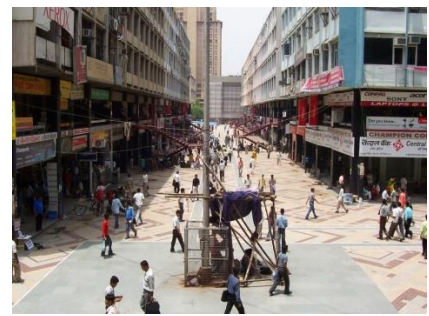

(a) Multiple Location     (b) Namesake False Location     (c) Too Small Font

**Fig. 14.** Represents the Drawbacks of Proposed Method